\documentclass{article}

\usepackage{PRIMEarxiv}

\usepackage[utf8]{inputenc} 
\usepackage[T1]{fontenc}    
\usepackage{hyperref}       
\usepackage{url}            
\usepackage{booktabs}       
\usepackage{amsfonts}       
\usepackage{nicefrac}       
\usepackage{microtype}      
\usepackage{lipsum}
\usepackage{fancyhdr}       
\usepackage{graphicx}       
\graphicspath{{media/}}     

\title{Variations of Squeeze and Excitation networks
}

\author{
  Mahendran NV\\
  \texttt{mhkannan5@gmail.com} \\
}

\begin{document}
\maketitle

\begin{abstract}
Convolutional neural networks learns spatial features and are heavily interlinked within kernels. The SE module have broken the traditional route of neural networks passing the entire result to next layer. Instead SE only passes important features to be learned with its squeeze and excitation (SE) module. We propose variations of the SE module which improvises the process of squeeze and excitation and enhances the performance. The proposed squeezing or exciting the layer makes it possible for having a smooth transition of layer weights. These proposed variations also retain the characteristics of SE module. The experimented results are carried out on residual networks and the results are tabulated.
\end{abstract}


\section{Introduction}
Deep neural network architectures are evolving since the winning of AlexNet \cite{alexnet,googlenet}. And from 2012, various neural networks architectures have been proposed \cite{resnet,efficientnet,convnet20,resnext,wrns,xception,mobilenet,mobilenet2,mobilenet3,vgg}. From then, it is believed that deeper architectures have been increasing the performance. Number of recent research developments reveals that other network characteristics like width, scale  also contribute to the performance \cite{efficientnet,resnet,resnext,wrns}. At present, depth is not made compulsory for network to perform better according to \cite{mobilenet}. Architectures proposed have various notable modules which is key for better learning, increased performance \cite{resnext,wrns}. Modules like Residual modules \cite{resnet} in resnets for weight degradation problem, Inception module \cite{googlenet} learning from different kernel reducing the computational cost to name a few paves way for creating a better performing network. Squeeze and excitation module in SE network \cite{sepaper} is different from the existing modules as it passes only the important information to the next layer.

Convolutional layer based networks work better for many visual based tasks like detection, classification \cite{alexnet,mobilenet}.  In the convolutional layer, kernel filters make the layer to learn features from the input. The features learned by the filters are spatially connected among the neighbourhood kernels and they are locally dependent. That is the reason, even regularization techniques like dropout fails to perform on CNNs \cite{dropblock}. Fully connected (FC) layer on the other hand are globally dependent. In most of the neural network architectures, Convolution layer is used from the initial stages of the network followed by FC layers. Convolutional layer learns all the features that are interdependent and then passes the results to the FC layers which learns globally dependent features.

In case of SE networks \cite{sepaper}, the SE modules contains FC layers and are added in between convolutional layers of the network. This phenomenon also makes input values of layer to squeeze and excite the result at the end. Linear fully connected layers passes the important global information as squeezing omits the non important information. Passing this information to the convolutional layer after excitation makes the layer to improve the representational power of the network. This is the core idea behind the squeeze and excitation network mechanism. Squeeze and excitation network performs feature recalibration over the network by squeezing the network features and bringing back to original form. This makes use of channels interdependencies in convolution to enhance the learning of network.

In this paper, we propose four types of SE module which utilizes the global dependencies to make the network generalize better. These variants have delayed squeeze or excite and also adds an additional convolutional layer in between SE module. Delaying the process of squeeze or excite makes the layer to pass better values as operation happens twice in delayed process. The information passed are much more concentrated when passing results the other layers. Similarly for other squeeze and excitation variants. And we also propose a new module called Bump module.

Bump module starts from the squeeze operation wherein the layer collects spatial information from the previous layers in a fully connected (FC) layer.The proposed method contains dense layer operations within the squeeze and excitation module. The global dependent features learned by the FC layers remain unutilized in the SE module. We implement FC operation after the linear which makes the layer to learn from global representations. This information is vital for the network to learn and generalize better. Rich consolidated information is passed to the FC layer of same size. This layer makes the network benefit from the vital information before reclaiming it to the older shape. This layer is followed by the linear layer of excitation part in the SE module. This proposed bump module not only passes the vital information but also makes the network learn from that information.

\begin{figure}
  \centering
  \includegraphics[width=\textwidth, height=8cm]{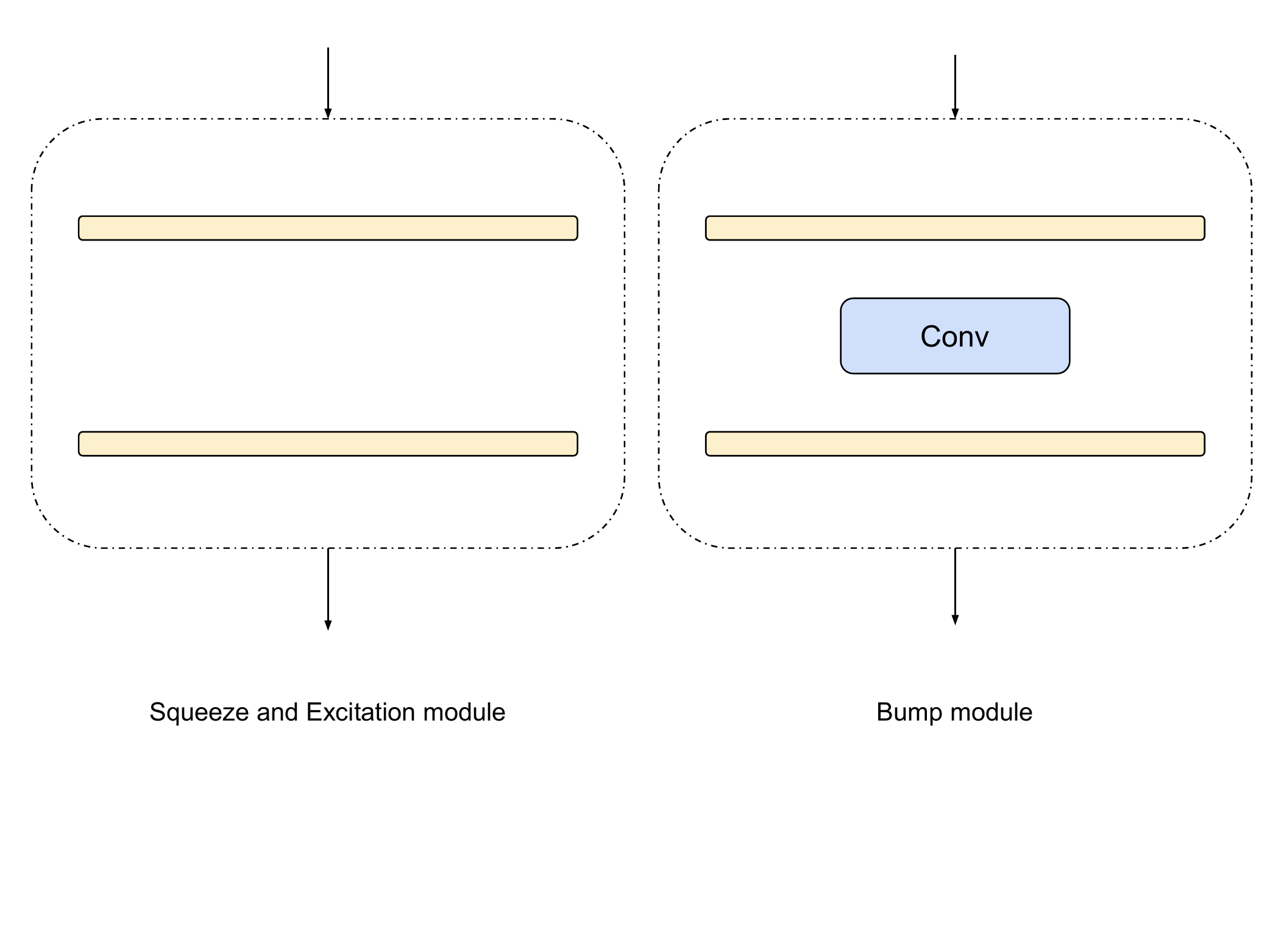}
  \caption{The comparison between squeeze and excitation module with the proposed module is shown in this figure. Bump module contains the dense layers on top and bottom similar to Squeeze and excitation module. Dense/FC layer on top gathers the input and passes it to the excitation method which makes the compressed values to regain its original shape for next layer}
  \label{fig:bump}
\end{figure}

We also shows the Bump module in detail in Figure \ref{fig:bump}. The main reason for calling it as bump module is that the new layer is added in between the SE operation which acts like a road bump in the module. This module is shown in comparison with the SE module. Bump module have single CNN layer that is placed in between squeeze and excitation layer as shown in Figure. In normal SE module, the excitation phase follows the squeeze layer of the network. Whereas in proposed network module, the convolution layer of size 3x3 is attached within squeeze to enhance performance of network.

\section{3-layer CNN}
\label{sec:headings}

We experiment the proposed method of learning in between squeeze and excitation in simple three layer CNN on CIFAR-10. We experiment the variety of proposed modules on three layer convolutional neural networks. We run the experiment with batch size of 64 and learning rate decayed with 0.7 with the starting value of 1.0. We use Adadelta optimizer with the stepwise learning rate. The experiment is run for 50 epochs.

\begin{figure}
  \centering
  \includegraphics[width=\textwidth, height=14cm]{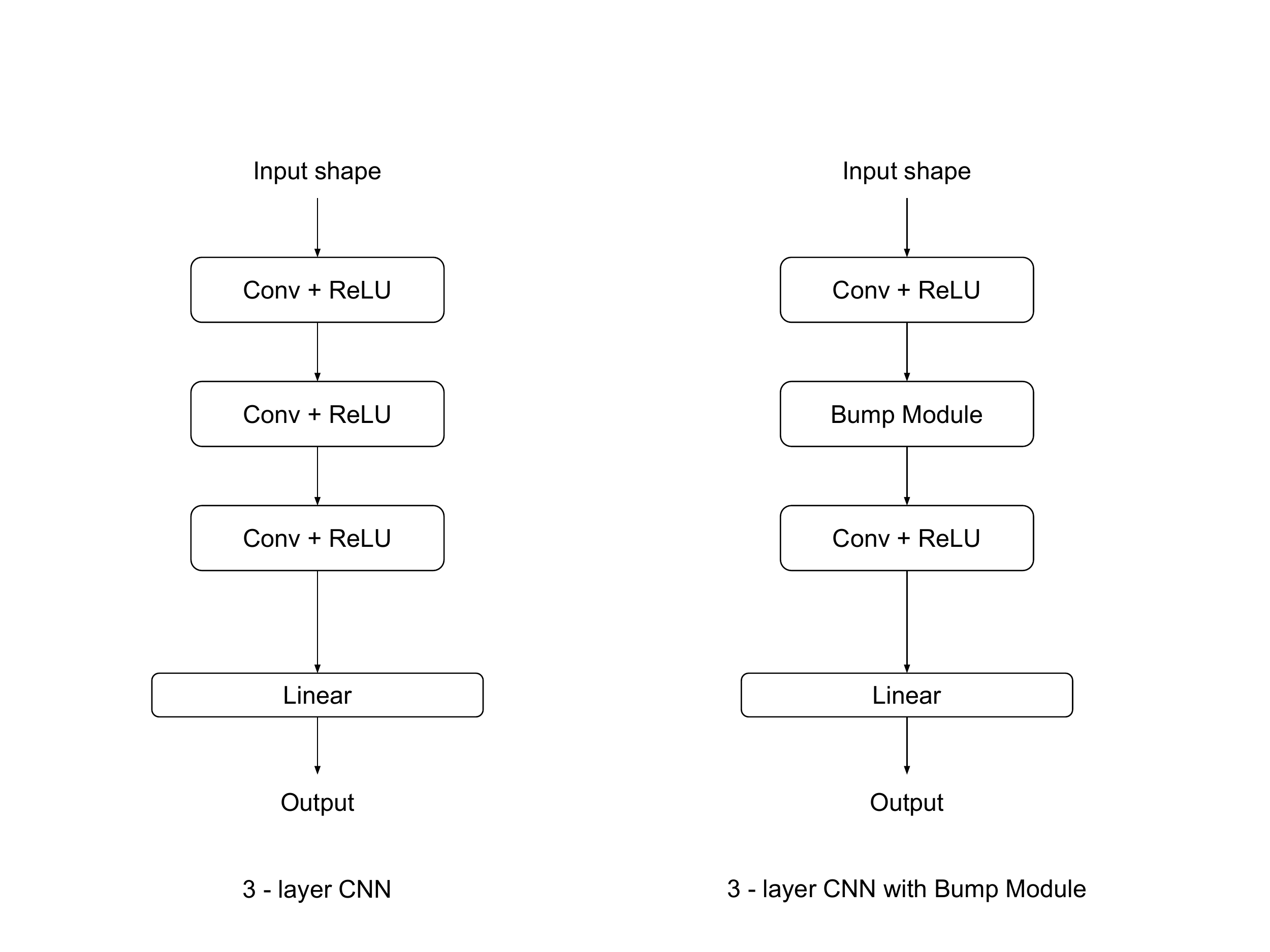}
  \caption{Three layer CNN. On the left: The default structure for the 3 layer cnn. On the right: The custom network structure with the proposed bump module with same configuration.}
  \label{fig:threecnn}
\end{figure}
The proposed modules are added on the three layer cnn and tested for performance. From the default 3 layers, if any module is needed to be tested, we replace the 2nd conv layer with the module and experiment on the CIFAR-10. For example, the bump module on 3 layer cnn is shown in figure \ref{fig:threecnn}.The results are tabulated as shown in \ref{tab:output_3cnn}. From the results, it is evident that the other variants of the proposed squeeze and excitation can enhance the performance in a better way. 

\begin{table}[]
    \centering
    \begin{tabular}{c|c|c}
        Method & Loss & Accuracy\\
        \hline
        Default & 0.8091 & 72.52 \\
        Squeeze and excite module & 0.8068 & 72.58 \\
        Slow Squeeze and excite & 0.8093 & 72.20 \\
        Squeeze and slow excite & 0.8131 & 72.63 \\
        Slow squeeze and slow excite & 0.7987 & 72.97 \\
        Bump module & 0.8044 & 72.65 \\
        \hline
    \end{tabular}

    \caption{Values represent the experiment carried out on 3-layer CNN with the default structure, existing SE module and the proposed variants.}
    \label{tab:output_3cnn}
\end{table}

There are other variants to the proposed approach. Existing alternatives are lowering the kernels sizes \cite{nin,sepaper}. There is another approach called gating mechanisms (highway) \cite{highway}. In some cases of using pretrained models, certain layers are freezed as inputs wont pass through while finetuning on dataset. Knowledge stored in those layers during training is not needed during fine tuning the model . This is somewhat similar to the gating mechanism. Some modules are also similar to this approach used in nn. Inception module, SE module, gating mechanisms and freezing layers (pretrained models).

\section{Literature review}
Deep learning networks: The interest in developing deep networks started after the winning of AlexNet \cite{alexnet} in ILSRVC 2012 competition. After the success, there are numerous architectures following the same path in providing deeper networks. There are notable architectures \cite{googlenet,efficientnet,mobilenet,resnet,wrns,resnext,sepaper} which harness the power of deep learning. The VGG \cite{vgg} network replaces larger filters in Alexnet with sequence of 3x3 filters won the ILSRVC 2014 competition. Residual networks make network to learn effectively while going deeper. The residual module have skip connections which passes the learned features after certain layers in turn solving vanishing gradient problem. One of the variants of resnet is ResNext which repeats the same block by having split-transform-merge strategy works better than ResNet. The ResNext block increases the width but proposed within the residual block which can be quite effective. Inception networks allows to have a sparse structure locally by introducing different kernel size convolutions in the inception module. Xception networks enhances faster training time by replacing the inception module with depthwise seperable convolutions.  Modules in state-of-the-art networks are improvised by introducing new techniques.

SqueezeNet uses a bottleneck approach to produce a small network. It consists of fire module which squeezes to certain planes then expand with kernel filters of 1 followed by 3 size. This model is smaller in size and passes the information effectively to next layer. Squeeze and excitation network proposes the s. Other variants of squeeze and then excite follows similar  approaches[a]. The core idea is to pass the important features when model is squeezed and excite in order to regain the original state of the network. Similar architectures have been proposed includes the network in networks which contains micro network in the network. This micronetwork consists of 1x1 filters traversing the network. As squeezenet paper claims this module capable of identifying non-linear dependencies in channels along with its model dynamic nature. This improves the representational power of network by passing important information. [Author] proposed highway networks which uses gating mechanism to regulate the information flow. Dualpath networks have the style of connecting paths internally. This network benefits from Resnet and Densenet capabilities and also learns new features with its internal connections among layers in turn increasing the performance of the network. Densely connected convolutional networks also have the redeveloping connections of layers which inherits similar capability of improving representational properties of the network.

Modules are improvised within and in case of squeeze methods in networks, the important information passed as output of squeeze method remains unutilized and directly regained to its original shape. The vital information is only trained after regaining its shape in the existing methods. There are other variations proposed to existing approaches for specific purposes \cite{convnet20,resnext,wrns,xception,mobilenet,mobilenet2,mobilenet3,variation3}. \cite{variation1} proposes the variations for alexnet \cite{alexnet} and googlenet \cite{googlenet} for having better performance on korean character recognition.

\section{Variations of SE}
This is the reason to add the variations in Squeeze and Excitation module in the network. This squeeze and excitation increases the representational power of the network. The variations are proposed in order to avoid sudden change in the network weights. SE module claims to pass only important information to the following layers. The proposed module tries to reduce the shapes into respective channels slowly so as to retain some of the unimportant values. The reason to keep the unimportant features is to make the layers avoid overfitting and learn better. [Explain it down with Experiment]

\begin{figure}
  \centering
  \includegraphics[width=\textwidth,height=16cm]{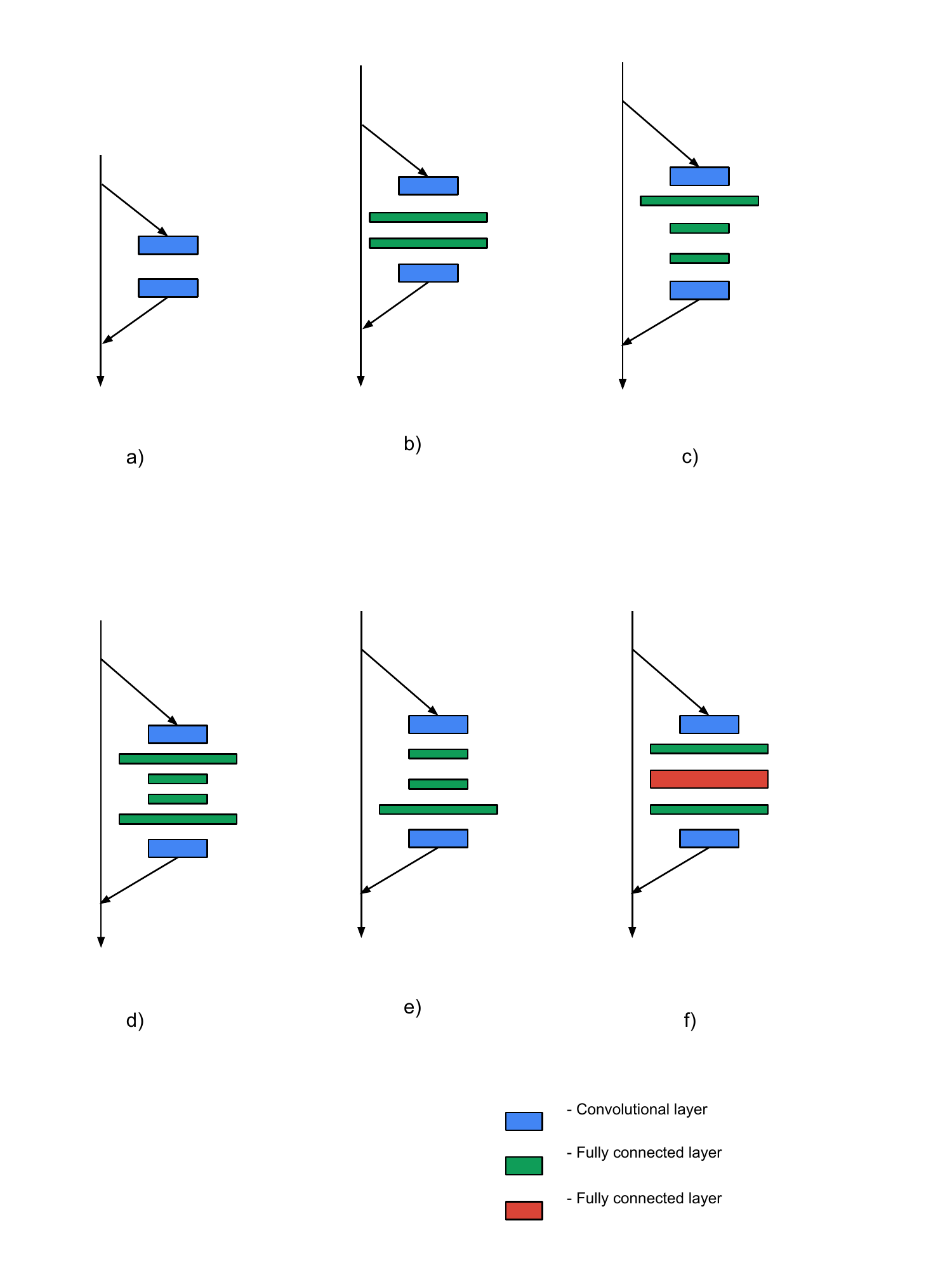}
  \caption{Visualizations represents the proposed variants of Squeeze and Excitation module \cite{sepaper} along with existing modules implemented on Residual module. a) Residual network module with two convolutional layers as used in Basicblock. b) The Squeeze and excitation module of SE \cite{sepaper} containing dense layers for squeeze and excitation. c) Slow squeeze and excite - Module makes the squeeze to delay by applying two dense layers for squeeze operation. d) Slow squeeze and slow excite - This variation have double fully connected layers for squeeze and excitation module. e) Squeeze and slow excite - This version contains slow excitation process in turn have two dense for excitation. f) This is new version of squeeze and excite which contains a working fully connected layer gets input in compressed squeeze layer and sends output to excitation layer for regaining original shape. }
  \label{fig:core}
\end{figure}

We propose four types of Squeeze and Excitation modules as shown in Figure. The SE module is compared with the four types of module as shown. Out of four proposed modules, three modules have differences in squeeze , excitation process. Remaining module have the FC layer within to enhance the performance. The types are explained in detail as follows

\textbf{Slow squeeze and excite} - In this module, the squeeze is reduced to two steps of squeeze. In the reduction process, the half of the size is reduced in the first, then the remaining half size of the input is reduced in the next step. The reduction value given is directly used while excitation step followed by squeeze.

\textbf{Slow squeeze and slow excite} - This module consists of changes in both squeeze and excitation processes. From the given reduction values, the squeeze is done for half of the reduction and passed to squeeze the remaining half. Similarly, in the excitation phase, the half of reduced output is excited and then the original shape is retained from the output of semi-excited shape.  

\textbf{Squeeze and slow excite} - In this module, the squeeze is completed for the given reduction shape whereas the excitation is slowed down a little. The squeezed result of (1x1xch) is excited to reduction//2 value. Then the output from this undergoes excitation of original reduction to retain the shape for passing to next stacked layers.

\textbf{Squeeze - learn - excite / Bump}- Unlike the previous methods, this model of squeeze and excitation consists of a fully connected layer in the middle. It is also called as Bump module as a FC layer acts as bump for default SE module. As researchers explained earlier [sepaper], the FC layer learns global representations and pass it on to next layers. In this module of Squeeze-learn-excite, the learned values are utilized directly in the squeeze state which is then excited to the normal state. 

\subsection{Training details}
We implement the proposed modules on Resnet-18. Residual networks makes the learning at particular instance and spread across nearby layers. We believe the residual network alone can help us show the experiment results in a better way. For the experiments we use CIFAR-10, CIFAR-100 and SVHN dataset. We use ResNet-18 as standard architecture for testing due to limited computational capabilities. We use the batch size of 32 with the datasets following normalization of its mean and standard deviation. We use three datasets and transformation for CIFAR-100 differs from the CIFAR-10. We use Random resized cropping of 224 image size for input data. The cropping is followed by horizontal flip in random manner. The training data also follows the normalization technique for CIFAR-100 dataset. For testing the CIFAR-100, the dataset is resized to 256 shape. Then the dataset undergoes center crop of 224 and then the normalization. For SVHN dataset, it follows similar approach to CIFAR-10 for both train and test dataset.

\section{Experimentation}
In this section, we will show the implementation of SE variants including Bump module on ResNet-18 network. Proposed variants on SE module outperforms other models with a large margin. The implementation and architecture details across models are added in this section. 

\section{Implementation}
Residual network is known for retaining the learned weights by omiting weight degradation problem when going deeper in networks. The proposed residual module consists of convolutional layer passes the learned parameters after certain layers. The learned parameters are relative to at that particular depth which makes it more effective as features that are passed just above certain layers are only used in that layer.

We implement the proposed variants slow squeeze and excite, slow squeeze and slow excite, squeeze and slow excite and the Bump module within the residual module. Since the residual module is possibly avoiding weight degradation, it is important to make residual module to pass vital information. The implementation of the variants including bump module on top of residual module is shown in the Figure \ref{fig:core}. We tested the proposed method on ResNet-18, with all additional latest techniques. The Batch Normalization technique is used which normalizes the output of convolutional layer batch by batch. This batch normalization is used widely as this is better than other normalization techniques. The results on the resnet-18 module for all the three datasets are tested and added in the Table.

\begin{table}[hbt!]
    \centering
    \begin{tabular}{ c | c | c }
        Methods & Top-1 & Top-5 \\
        Default & 73.130 & 97.620\\
        SE module & 75.580 & 97.210 \\
        Slow squeeze and excite & 74.370 & 96.910 \\
        Squeeze and slow excite & 75.630 & 97.160 \\
        Slow squeeze and slow excite & 76.020 & 97.700 \\
        Bump module & 76.020 & 97.810 \\
        \vspace{0.3cm}
    \end{tabular}
    \caption{Results tested on CIFAR-10 dataset on the Resnet-18 with proposed variants of SE module.}
    \label{tab:res_cifar10}
\end{table}

\subsection{CIFAR - 100 results}

\begin{table}[hbt!]
    \centering
    \begin{tabular}{ c | c | c }
        Method & Top-1 & Top-5 \\
        Default & 72.870 & 92.380 \\
        SE module & 74.460 & 93.190 \\
        Slow squeeze and excite & 73.340 & 92.960 \\
        Squeeze and slow excite & 73.500 & 93.910 \\
        Slow squeeze and slow excite & 73.250 & 92.860 \\
        Bump module & 73.960 & 93.200 \\
    \end{tabular}
    \vspace{0.3cm}
    \caption{Results tested on CIFAR-100 dataset on the Resnet-18 with proposed variants of SE module.}
    \label{tab:res_cifar100}
\end{table}
Table \ref{tab:res_cifar100} represents the results with above mentioned configurations. The values prove that the proposed module does have effect over basic network. The results are on par with SE module in case of top-5 accuracies and sometimes its even better than SE module.
\subsection{SVHN results}

\begin{table}[hbt!]
    \centering
    \begin{tabular}{ c | c | c }
        Method & Loss & Accuracy \\
        Default & 91.211 & 98.851 \\
        SE module & 91.795 & 98.855 \\
        Slow squeeze and excite & 92.048 & 98.959 \\
        Squeeze and slow excite & 92.244 & 98.871 \\
        Slow squeeze and slow excite & 92.187 & 98.782 \\
        Bump module & 91.576 & 98.882 \\
    \end{tabular}
    \vspace{0.3cm}
    \caption{Results tested on SVHN dataset on the Resnet-18 with proposed variants of SE module.}
    \label{tab:res_svhn}
\end{table}

From the results, the proposed modules enhances the performance of model upto 1.01 percentage in case of SVHN dataset. This enhancement comes as every residual layer makes sure it passes only key features of the image to the network.

\section{Conclusion}
We propose variations of squeeze and excitation module on residual network. The variations on residual makes the network to pass vital information. The proposed SE variants can be applied for the existing networks similar to SE module. We presented the variants for Squeeze and excitation, which is similar to Mobilenetv2, Mobilenetv3 and like Inceptionv2,inceptionv3. We expect the Squeeze and excitation and its variants, a cornerstone for traditional input flow over network as it passes only vital information.

\bibliographystyle{unsrt}  
\bibliography{references}  

\end{document}